%% file: main.tex
\documentclass[conference]{IEEEtran}

\usepackage{graphicx}
\usepackage{amsmath,amssymb,amsfonts}
\usepackage[T1]{fontenc}
\usepackage{url}
\usepackage{hyperref}
\usepackage{xcolor}
\usepackage{siunitx}
\usepackage[linesnumbered,ruled,vlined]{algorithm2e} 
\usepackage{algpseudocode}
\usepackage{caption}
\usepackage{float}
\usepackage{subcaption}
\usepackage[section]{placeins}
\usepackage{stfloats}
\usepackage{svg}
\usepackage{booktabs}
\usepackage{multirow}
\usepackage{graphicx}
\usepackage{adjustbox}
\usepackage{tabularx,booktabs,ragged2e}
\usepackage{amsmath,amssymb}
\usepackage{enumitem}
\usepackage{tcolorbox}
\usepackage[ruled,vlined]{algorithm2e}
\usepackage[ruled,vlined,linesnumbered]{algorithm2e}

\usepackage{listings}
\usepackage{tabularx}          
\newcolumntype{L}[1]{>{\raggedright\arraybackslash}p{#1}}

\begin{document}


\title{SynLLM: A Comparative Analysis of Large Language Models for Medical Tabular Synthetic Data Generation via Prompt Engineering}

\author{
\IEEEauthorblockN{Arshia Ilaty\IEEEauthorrefmark{1},
Hossein Shirazi\IEEEauthorrefmark{2},
Hajar Homayouni\IEEEauthorrefmark{2}}
\IEEEauthorblockA{\IEEEauthorrefmark{1}Computational Science Research Center, San Diego State University}
\IEEEauthorblockA{\IEEEauthorrefmark{2}Department of Computer Science, San Diego State University\\
Email: \{ailaty, hshirazi, hhomayouni\}@sdsu.edu}
}


\maketitle
\begin{abstract}

\input{00-Abstract}

\end{abstract}


\section{Introduction}
\input{10-Introduction}

\section{Related Work}
\label{sec:related_work}
\input{20-RelatedWork}

\section{Methodology}
\label{sec:methodology}
\input{30-Methodology}

\section{Results and Analysis}
\label{sec:results}

\input{40-Results}

\section{Key Observations and Discussion}
\label{sec:discussion}

\input{50-Discussion}

\section{Conclusion and Future Directions}
\label{sec:future_work}
\input{60-Conclusion}


\bibliographystyle{plain}
\bibliography{references}

\clearpage
\section{Appendix}
\label{sec:supplementary}
\input{70-Supplementary}

\end{document}

%% file: 00-Abstract.tex
Access to real-world medical data is often restricted due to privacy regulations, posing a significant barrier to the advancement of healthcare research. Synthetic data offers a promising alternative; however, generating realistic, clinically valid, and privacy-conscious records remains a major challenge. Recent advancements in Large Language Models (LLMs) offer new opportunities for structured data generation; however, existing approaches frequently lack systematic prompting strategies and comprehensive, multi-dimensional evaluation frameworks.

In this paper, we present \textbf{SynLLM}, a modular framework for generating high-quality synthetic medical tabular data using 20 state-of-the-art open-source LLMs, including LLaMA, Mistral, and GPT variants, guided by structured prompts. We propose four distinct prompt types, ranging from example-driven to rule-based constraints, that encode schema, metadata, and domain knowledge to control generation without model fine-tuning. Our framework features a comprehensive evaluation pipeline that rigorously assesses generated data across statistical fidelity, clinical consistency, and privacy preservation.

We evaluate SynLLM across three public medical datasets, including \textit{Diabetes}, \textit{Cirrhosis}, and \textit{Stroke}, using 20 open-source LLMs. Our results show that prompt engineering significantly impacts data quality and privacy risk, with rule-based prompts achieving the best privacy-quality balance. SynLLM establishes that, when guided by well-designed prompts and evaluated with robust, multi-metric criteria, LLMs can generate synthetic medical data that is both clinically plausible and privacy-aware, paving the way for safer and more effective data sharing in healthcare research.

\begin{IEEEkeywords}
Synthetic Data Generation, Large Language Models, Tabular Medical Data, Privacy, Prompt Engineering, Healthcare AI
\end{IEEEkeywords}

%% file: 10-Introduction.tex
Access to real-world medical data is frequently restricted due to privacy regulations, ethical constraints, and institutional barriers, posing a significant challenge for the development of AI-driven healthcare solutions. While data protection laws such as the Health Insurance Portability and Accountability Act (HIPAA)~\cite{hipaa1996} and the General Data Protection Regulation (GDPR)~\cite{gdpr2016} are essential for safeguarding patient confidentiality, they often hinder the availability of data for clinical model development and research. Synthetic data offers a promising alternative by enabling the training and validation of machine learning models without exposing real patient records.

Existing approaches to structured synthetic data generation, including Generative Adversarial Networks (GANs), Variational Autoencoders (VAEs), and more recently, Large Language Models (LLMs), have shown potential but suffer from key limitations. GAN-based methods like CTGAN~\cite{ctgan} and MedGAN~\cite{macedo2024medgan} frequently experience mode collapse and require large amounts of real training data, limiting their utility in privacy-sensitive contexts~\cite{goodfellow2014generative}. VAEs tend to oversmooth feature distributions, thereby suppressing rare but clinically important conditions~\cite{vae_medical}. Additionally, both GANs and VAEs often struggle to capture complex feature interdependencies, resulting in synthetic records that lack medical plausibility.

Recent advancements in LLMs, including GReaT \cite{borisov2023languagemodelsrealistictabular}, and REaLTabFormer \cite{solatorio2023realtabformer}, present new opportunities for generating high-quality and privacy-preserving structured synthetic data. When guided with structured prompts, LLMs can produce contextually rich and statistically aligned tabular data. However, current LLM-based approaches face critical challenges:

\noindent \textbf{Lack of structured prompting.} Most existing methods rely on unstructured text generation followed by post-processing to construct tabular data, which is additional overhead and can introduce errors.

\noindent \textbf{Privacy risks.} Without explicit and effective design constraints, LLMs may memorize and inadvertently replicate sensitive training records.

\noindent \textbf{Research Goals.} This work aims to investigate how prompt structure affects the quality and privacy of LLM-generated synthetic medical data. Specifically, we (1) develop a set of prompt strategies that encode schema information, statistical metadata, and clinical logic; (2) evaluate the ability of open-source LLMs to generate realistic and privacy-preserving synthetic records under these prompts; and (3) quantify performance trade-offs using a multidimensional evaluation framework that spans statistical fidelity, medical plausibility, and privacy risk. 

\noindent \textbf{Proposed Approach.} To study how prompt structure affects synthetic data generation, we introduce \textbf{SynLLM}, a prompt-driven evaluation framework for structured medical data synthesis using LLMs. SynLLM implements four systematically designed prompt types, ranging from minimal information prompts that provide only column headers and a few example records to metadata-augmented and rule-based prompts that incorporate statistical summaries and domain-specific clinical constraints. Notably, the final prompt type excludes all example records and relies solely on rule-based guidance, allowing us to evaluate model performance under stricter privacy-aware generation conditions. These prompts guide LLMs in generating structured tabular records without requiring model fine-tuning. This design enables controlled comparisons of prompt effectiveness and supports the analysis of how different prompting strategies influence data quality, clinical validity, and privacy risk.

\noindent \textbf{Evaluation and Findings.} SynLLM is evaluated across three public medical datasets—\textit{Diabetes}, \textit{Cirrhosis}, and \textit{Stroke} using 20 open-source LLMs, including Mistral-7B, Zephyr-7B, LLaMA, and GPT-2. Results demonstrate that prompt structure significantly impacts output quality and privacy. Rule-based prompts consistently achieve high harmonic privacy-quality scores without relying on example records. Our evaluation reveals that model behavior varies substantially across prompt types, highlighting the importance of prompt design in LLM-guided synthetic data generation.

\noindent \textbf{Structure of the Paper.} Section~\ref{sec:related_work} reviews relevant literature in synthetic data generation. Section~\ref{sec:methodology} introduces the SynLLM pipeline and prompt types. Section~\ref{sec:evaluation} presents experimental setup and evaluation metrics. Section~\ref{sec:results} provides empirical results and analysis. Section~\ref{sec:discussion} provides key observations, followed by conclusions and future directions in Section~\ref{sec:future_work}.

%% file: 20-RelatedWork.tex
The generation of synthetic medical data has been explored through a variety of modeling paradigms, including traditional generative models, privacy-preserving algorithms, and, more recently, large language models (LLMs). This section surveys the landscape of existing approaches, highlighting their contributions and limitations in the context of fidelity and privacy.

We first review LLM-based frameworks that utilize transformer architectures for tabular data generation. Next, we summarize alternative generative methods such as GANs, VAEs, and diffusion models, which have been widely adopted in synthetic tabular data research. Finally, we discuss techniques that explicitly incorporate privacy preservation through mechanisms such as differential privacy or post hoc filtering. We conclude the section by situating SynLLM within this landscape and explaining how it addresses limitations identified in prior work.

\subsection{LLM-Based Approaches for Synthetic Medical Data}
Recent advancements in Large Language Models (LLMs) have demonstrated their ability to generate structured medical data by capturing complex feature interdependencies. GReaT introduced text-based encoding for tabular records, improving data diversity; however, with computational overhead and privacy risks. HARMONIC \cite{Wang2024HARMONICHL} presented instruction-tuned LLMs with $k$-nearest neighbors strategies that improved privacy preservation, though its evaluation metrics lack granularity in detecting structured privacy violations. 



\subsection{Alternative Generative Models}
Traditional models such as GANs (medGAN) and CTGAN improved categorical variable handling but suffer from mode collapse, computational intensity, and training sensitivity. VAEs provide smooth latent representations but generate overly averaged data, missing rare but critical cases. Diffusion models like TabDDPM \cite{kotelnikov2023tabddpm} enhance distributional accuracy but require extensive computational resources.

\subsection{Privacy-Preserving Approaches}
Privacy-preserving techniques include DP-integrated methods, including DP-SDG \cite{PHONG2023102819}, DP-GAN \cite{HO2021103066}, and DP-WGAN \cite{9993791} that inject noise into training procedures but often degrade synthetic data utility. Recent DP-enhanced LLM models like DP-LLMTabGen \cite{dpllmtabgen} show promise in balancing privacy and statistical fidelity.
In contrast, our proposed SynLLM framework addresses these limitations through structured prompt engineering that embeds clinical logic and statistical properties explicitly. This approach maintains medical coherence, reduces computational overhead, and eliminates the need for latent-space modeling while enforcing metadata properties and domain-specific rules at generation time. SynLLM provides greater flexibility through prompt-based generation without requiring model retraining for different subpopulations.

%% file: 30-Methodology.tex
This section outlines the design and components of the SynLLM framework for structured synthetic medical data generation using LLMs. SynLLM is built around a modular pipeline that includes schema profiling, prompt construction, LLM-based record generation, and multi-dimensional evaluation. The core methodological innovation lies in the use of structured, domain-informed prompts that guide generation without requiring model retraining or fine-tuning. We describe the four prompt strategies employed, the data generation process across 20 open-source LLMs, and the multi-dimensional evaluation criteria used to assess statistical fidelity, clinical consistency, privacy preservation, and computational efficiency. The following subsections detail each stage of the pipeline.

\subsection{Problem Definition}\label{sec:probdef}

Let $\mathcal{D}_{\text{real}} = \{(x^{(i)}, y^{(i)})\}_{i=1}^N$ denote a structured electronic health record (EHR) dataset, where each row $x^{(i)} \in \mathbb{R}^{p_{\text{num}}} \times \mathcal{C}^{p_{\text{cat}}}$ comprises $p_{\text{num}}$ numerical and $p_{\text{cat}}$ categorical attributes, and $y^{(i)}$ is an optional downstream label.

We define a prompt-driven generation mechanism
\[
\mathcal{G}_\theta : (\Pi,\,k) \longmapsto \hat{\mathcal{D}}_{\text{syn}}
\]
that, given a prompt specification $\Pi$ and a target record count $k \ll N$, produces a synthetic dataset $\hat{\mathcal{D}}_{\text{syn}}$ such that:

\begin{enumerate}[leftmargin=*]
    \item \textbf{Statistical fidelity:} $\hat{\mathcal{D}}_{\text{syn}}$ approximates the marginal and joint distributions of $\mathcal{D}_{\text{real}}$ within a tolerance $\varepsilon_{\text{stat}}$.
    
    \item \textbf{Clinical plausibility:} Synthetic records satisfy logical and medical constraints (e.g., \texttt{HbA1c} $>$ 6.5 $\Rightarrow$ \texttt{Diabetes} = True).
    
    \item \textbf{Privacy preservation:} The probability that any $\hat{x} \in \hat{\mathcal{D}}_{\text{syn}}$ is linkable to a real record is bounded above by $\delta_{\text{priv}}$, as estimated via empirical privacy metrics (e.g., $k$-anonymity, membership inference, nearest-neighbor distance).
\end{enumerate}

Unlike GAN- or VAE-based methods, which require access to real patient records during model training, SynLLM leverages zero- and few-shot LLM inference guided by carefully designed prompts. These prompts incorporate only aggregate statistics and domain rules extracted from $\mathcal{D}_{\text{real}}$, without exposing any individual-level data. By operating exclusively on non-identifiable summaries, including feature distributions, clinical thresholds, and correlation patterns, SynLLM reduces disclosure risk while exploiting the rich prior knowledge encoded in modern instruction-tuned language models~\cite{borisov2023languagemodelsrealistictabular,solatorio2023realtabformer}.


\subsection{SynLLM Framework Overview}\label{sec:pipeline}

The SynLLM pipeline (Algorithm~\ref{alg:synllm}) consists of four modular stages that enable LLM-based generation of privacy-conscious, clinically meaningful structured medical data. Each stage is designed to preserve fidelity to real data characteristics while minimizing privacy risks. An overview is as follows:

\begin{enumerate}[leftmargin=*]
    \item \textbf{Schema Analysis.} Extract attribute types, univariate statistics, and relevant inter-feature correlations from $\mathcal{D}_{\text{real}}$ (Sec.~\ref{sec:schema}). Only aggregated metadata, never raw records, are surfaced outside the secure data enclave.

    \item \textbf{Prompt Construction.} Construct a generation prompt $\Pi$ using one of four progressively constrained templates, each encoding different levels of statistical metadata and clinical logic (Sec.~\ref{sec:prompts}).

    \item \textbf{LLM Inference.} Query an instruction-tuned, open-source language model (see Table~\ref{tab:models}) using fixed sampling parameters (temperature $T{=}0.7$, top-$p{=}0.9$). The token budget is dynamically adjusted based on the desired record count $k$.

    \item \textbf{Post-processing and Validation.} Parse generated JSON objects into structured tabular form, enforce data typing constraints, and discard records violating hard-coded clinical rules. Validated records are passed to the evaluation pipeline described in Sec.~\ref{sec:results}.
\end{enumerate}

\begin{algorithm}[ht]
\caption{SynLLM: Structured Medical Data Generation with LLMs}
\label{alg:synllm}
\KwIn{Real dataset ${\mathcal{D}}_{\text{real}}$ (for schema extraction only), set of LLMs $\mathcal{M}$, prompt templates $\mathcal{P}$}
\KwOut{Synthetic dataset $\hat{\mathcal{D}}_{\text{syn}}$ with statistical, clinical, and privacy evaluations}

\BlankLine
\textbf{Stage 1: Metadata Extraction}\\
Extract feature schema $\mathcal{S}$, value ranges, types, and statistical summaries from ${\mathcal{D}}_{\text{real}}$\;
Identify domain rules and clinical constraints $\mathcal{R}$ from medical knowledge base or expert guidance\;

\BlankLine
\textbf{Stage 2: Prompt Engineering}\\
\ForEach{prompt type $p \in \mathcal{P}$}{
    Construct prompt $P$ using schema $\mathcal{S}$, metadata, and rules $\mathcal{R}$\;
}

\BlankLine
\textbf{Stage 3: Synthetic Data Generation}\\
\ForEach{model $m \in \mathcal{M}$}{
    \ForEach{prompt $P$}{
        Generate synthetic records $R_{m,P} = m(P)$\;
        Parse $R_{m,P}$ into structured tabular format\;
    }
}

\BlankLine
\textbf{Stage 4: Evaluation and Filtering}\\
\ForEach{synthetic record set $R_{m,P}$}{
    Compute statistical metrics (e.g., Wasserstein, correlation)\;
    Compute medical consistency scores based on $\mathcal{R}$\;
    Compute privacy risk metrics (e.g., k-anonymity, NN distance)\;
    Optionally filter or flag low-quality or high-risk records\;
}
\Return{$\hat{\mathcal{D}}_{\text{syn}} = \bigcup R_{m,P}$}
\end{algorithm}
\setlength{\textfloatsep}{10pt}
\subsection{Adaptive Prompt Taxonomy}\label{sec:prompts}
Our prompt schema is organized into a four‑tier hierarchy of escalating sophistication: Level 1 functions as the baseline, while levels 2 through 4 incrementally introduce richer contextual cues, including feature definition and statistical properties, and stricter domain‑specific constraints.

\begin{description}[leftmargin=0cm,labelsep=0.2cm,style=sameline,font=\bfseries]
\item[\textsc{SeedEx} (Prompt‑A):] \textit{Example‑Seed Minimal Prompt}.  
Lists the column headers corresponding to dataset features, the desired output format, and $\leq5$ seed rows randomly sampled from $\mathcal{D}_{\text{real}}$.  
Purpose: to establish a \emph{baseline} that stresses model generalization under minimal constraint. However, this formulation presents the highest risk of record memorization and identity disclosure.

\item[\textsc{FeatDesc} (Prompt‑B):] \textit{Feature‑Description Prompt}.  
Replaces concrete examples with concise natural‑language definitions of each attribute (e.g.\ “\texttt{bmi}: body‑mass index in kg/m$^2$, a continuous variable bounded within $[12, 60]$).  
This approach introduces semantic structure by providing the model with descriptive, clinically grounded definitions of each feature, which guide the generation process and help constrain outputs to realistic, in-distribution value ranges.

\item[\textsc{StatGuide} (Prompt‑C):] \textit{Statistical‑Metadata Prompt}.  
Extends \textsc{FeatDesc} with feature-level summaries including means, standard deviations, min–max bounds, category frequencies, and selected pairwise correlations. 
This template draws inspiration from the “data portrait” concept in \cite{Wang2024HARMONICHL}, which encodes statistical summaries to guide generation. In our framework, we apply similar dataset-level metadata to construct the \textsc{StatGuide} prompt, which has been empirically shown to reduce divergence from the target distribution (Sec.~\ref{sec:results}).

\item[\textsc{ClinRule} (Prompt–D):] \textit{Clinically-Constrained Prompt}.  
Eliminates example records entirely and replaces them with declarative logic rules derived from medical guidelines (e.g., “If \texttt{pregnant}=True, then \texttt{sex}=Female”).  
The LLM is required to generate samples that satisfy these constraints, thereby prioritizing logical consistency and minimizing disclosure risk.
\end{description}

Table~\ref{tab:prompt-excerpt} provides an abridged overview of each prompt template. All prompts share a consistent \textbf{system message} instructing the model to (i) emit \texttt{newline}-delimited JSON objects, (ii) avoid free-text commentary, (iii) adhere to the requested number of records, and (iv) refrain from emitting any protected health information (\textit{PHI}). During prompt construction, dataset-specific metadata is programmatically inserted into placeholder tags (e.g., \texttt{\{feature\_stats\}}).

\begin{table}[t]
\centering\small
\caption{Prompt skeletons (abridged). Curly braces denote runtime placeholders.}
\label{tab:prompt-excerpt}
\begin{tabularx}{\columnwidth}{@{}L{0.24\columnwidth}X@{}}
\toprule
\textbf{Template} & \textbf{Key Sections}\\
\midrule
\textsc{SeedEx}   & Header row; $n$ example records; ``Repeat format exactly, $k$ rows.''\\
\textsc{FeatDesc} & Header row; per-feature descriptions; JSON schema block.\\
\textsc{StatGuide}& As \textsc{FeatDesc}, plus \{mean\}, \{stdev\}, \{min,max\}, frequency tables; optional correlation matrix snippet.\\
\textsc{ClinRule} & Header row; domain-specific logic rules (e.g., DL $\rightarrow$ HbA1c $>$ 6.5); JSON schema; no examples.\\
\bottomrule
\end{tabularx}
\end{table}

\paragraph*{Design Rationale}
This prompt taxonomy systematically varies the amount and type of conditioning information supplied to the LLM, allowing for controlled exploration of the privacy–utility trade-off. \textsc{SeedEx} provides minimal constraint, often resulting in low Jensen–Shannon divergence but elevated membership inference risk. At the opposite end, \textsc{ClinRule} imposes strict domain rules, substantially mitigating privacy risk at the expense of greater distributional shift. The intermediate templates—\textsc{FeatDesc} and \textsc{StatGuide}—introduce semantic and statistical context, enabling precise evaluation of how information content affects fidelity and generalization. Empirical results in Sec.~\ref{sec:results} show that \textsc{StatGuide} achieves the best utility for internal analytics, while \textsc{ClinRule} is most suitable for public release scenarios.

\paragraph*{Schema and Statistical Extraction}\label{sec:schema}
For each numerical attribute $f$, we extract the 5-tuple $(\mu_f, \sigma_f, \text{min}_f, \text{max}_f, \text{quantiles}_f)$. For categorical attributes, we compute the empirical probability mass function $\mathbf{p}_f$. To reduce the risk of rare-category disclosure, we apply a frequency threshold of five and consolidate infrequent values into an “\texttt{Other}” category before incorporating them into prompt metadata. Pairwise Pearson correlations $\rho_{fg}$ are retained only if $|\rho_{fg}| > 0.15$ or identified as clinically relevant by domain experts.

\subsection {Evaluation and Metrics Description}
\label{sec:evaluation}

To evaluate the effectiveness of the SynLLM framework, we conduct a comprehensive \textbf{quality–privacy–utility audit} that assesses each synthetic dataset across four orthogonal performance dimensions:

\begin{enumerate}[label=\textbf{\arabic*.},leftmargin=*]
    \item \textbf{Statistical fidelity} — We assess marginal and joint distribution alignment using metrics that include the Kolmogorov-Smirnov, $\chi^2$ and the Wasserstein distance. Thresholds are applied to flag significant divergence between real and synthetic data.

    \item \textbf{Clinical consistency} — A rule engine based on evidence-informed medical constraints (e.g., ADA, WHO) validates generated records against known physiological and logical dependencies. Records that violate hard constraints (e.g., biologically implausible values or contradictory labels) are flagged or discarded.

    \item \textbf{Privacy protection} — We evaluate disclosure risk using empirical, distance-based metrics. Specifically, we compute nearest-neighbor distance ratios and identifiability scores to estimate the likelihood that synthetic records closely resemble real ones. Synthetic datasets are flagged if these privacy metrics fall below a pre-specified threshold $\delta_{\text{priv}}$.

    \item \textbf{Machine learning utility} — Tree-based classifiers (e.g., decision tree, random forest, XGBoost) are trained and evaluated under both TSTR and TRTS paradigms. Synthetic datasets are retained only if performance gaps in accuracy, macro-F1, or AUC-ROC remain within an acceptable range $\varepsilon_{\text{util}}$ compared to real-data baselines.
\end{enumerate}


\subsubsection{Statistical Fidelity Assessment}

To ensure that synthetic data generated by SynLLM faithfully mirrors the structure of the original dataset, we evaluate \textbf{statistical fidelity} using a targeted set of distributional and relational metrics.
These are designed to capture alignment in marginal distributions, pairwise dependencies, and categorical structure, each essential to preserving the analytical and statistical utility of medical data.
For each to these areas, we collected metrics and measured those metrics in our experiments. In the following, we explained list of these metrics for each group.

\vspace{0.5em}
\noindent \textbf{Marginal Distribution Alignment.}  
To evaluate whether the generated features follow the same value distributions as the real data, we apply:

\textbf{Wasserstein Distance}~\cite{villani2009optimal} — Quantifies the cost of morphing one distribution into another, suitable for comparing empirical numerical distributions.

\textbf{Jensen–Shannon Divergence}~\cite{menendez1997jensen} — A bounded, symmetric divergence metric that is robust to support mismatches.

\textbf{Anderson–Darling k-Sample Test}~\cite{scholz1987k} — Detects differences between distributions with enhanced sensitivity in the tails.

\textbf{Kullback–Leibler Divergence}~\cite{lin1991divergence} — Measures information loss when approximating real data with synthetic estimates.

\textbf{Range Coverage} — Computes the proportion of the real-valued range covered by the synthetic data for each numerical feature. This detects both undercoverage (missing extreme cases) and overcoverage (hallucinated or out-of-distribution values).

\vspace{0.5em}
\noindent \textbf{Dependency and Correlation Preservation.}  
To assess whether inter-feature relationships are preserved, a key requirement for clinical realism, we compute:

\textbf{Pearson Correlation Coefficients}~\cite{pearson1895correlation} — Evaluate linear dependencies between features.

\textbf{Frobenius Norm of Correlation Matrix Differences}~\cite{golub2013matrix} — Captures global structural deviation in correlation networks.

\textbf{Feature-Level Correlation Analysis}~\cite{xu2019modeling} — Inspects preservation of specific medically relevant relationships (e.g., age vs. glucose).

\vspace{0.5em}
\noindent \textbf{Categorical Structure Fidelity.}  
To validate whether category distributions are retained, especially for rare conditions, we apply:

\textbf{$\chi^2$ Test and p-values}~\cite{pearson1900x} — Compare category frequency distributions.

\textbf{Category Preservation Rate}~\cite{choi2017generating} — Measures how well the diversity of categorical values is retained.

\textbf{Mutual Information Score}~\cite{cover1999elements} — Captures co-dependence among categorical variables, important for diagnosis-treatment modeling.

Together, these metrics allow us to quantify fidelity from three complementary angles: how realistic each feature’s distribution is, how well statistical dependencies are preserved, and whether categorical structure remains intact. This triangulated approach provides robust support for downstream analytics, risk modeling, and simulation tasks.

\subsubsection{Clinical Consistency Evaluation}

While statistical similarity is a necessary condition for synthetic data quality, it is not sufficient for clinical relevance. To ensure that synthetic records preserve medically meaningful relationships, we evaluate \textbf{clinical consistency} using a set of domain-informed metrics grounded in epidemiological and physiological principles.
These metrics are selected based on known risk factors and clinical patterns relevant to the datasets used in our study (e.g., Diabetes and Stroke). They assess whether key associations between features, including disease status, demographics, and laboratory results, are preserved in the synthetic cohort.

\vspace{0.5em}
\noindent \textbf{Dataset-Specific Examples.}  
For illustrative purposes, we include the following checks:

\textbullet{\textbf{HbA1c level differences}} between diabetic and non-diabetic subgroups \\
\textbullet{ \textbf{Mean glucose levels}} stratified by stroke outcome \\
\textbullet{ \textbf{Age-based stroke risk gradients}} consistent with clinical trends \\
\textbullet{ \textbf{Hypertension–stroke co-occurrence patterns}}, reflecting expected comorbidities
Each comparison is computed using group-wise mean differences or deviations in regression slopes relative to the real dataset.

\vspace{0.5em}
\noindent \textbf{Aggregation and Interpretation.}  
The deviations are aggregated into a \textit{clinical consistency score}, where lower values indicate closer alignment with expected clinical patterns. This score helps identify models or prompts that generate semantically plausible but medically inconsistent outputs.

\vspace{0.5em}
\noindent While this evaluation is not exhaustive across all possible clinical scenarios, it provides targeted validation of whether high-level medical logic is preserved in synthetic data generated under diverse prompt-model configurations.

\subsubsection{Privacy Risk Evaluation}

SynLLM assesses privacy risk using empirical, distance-based metrics commonly adopted in synthetic data literature. These metrics estimate the likelihood that synthetic records closely resemble or directly replicate real individuals in the source dataset, without enforcing formal privacy guarantees.

\noindent \textbf{Nearest Neighbor Distance Ratio.} For each synthetic record, we compute the Euclidean distance to its closest match in the real dataset, and compare this to the average nearest-neighbor distance among real records. The resulting privacy score is defined as the ratio of these averages. Higher values indicate stronger privacy, as synthetic records remain well-separated from real ones.

\noindent \textbf{Identifiability Score.} We also compute the fraction of synthetic records that are exact duplicates of records in the real dataset (i.e., identical across all features). Lower values are preferable, as they reflect reduced risk of direct leakage or memorization.

These distance-based metrics provide interpretable, model-agnostic signals of potential disclosure risk. However, SynLLM does not implement formal differential privacy guarantees, $k$-anonymity, or adversarial membership inference attacks. As such, this assessment should be understood as an empirical audit rather than a formal privacy certification.

Synthetic data batches that exhibit high privacy risk scores or violate anonymity thresholds are logged for further analysis and may inform prompt refinement or post-processing strategies in subsequent iterations of the generation pipeline.

\subsubsection{Machine Learning Utility}

In addition to statistical and clinical alignment, synthetic data must support real-world downstream tasks. We evaluate \textbf{machine learning utility} by assessing whether models trained on synthetic data yield predictive performance comparable to those trained on real data. This analysis ensures that SynLLM-generated data preserves not only feature-level distributions but also task-relevant signal for classification, without compromising privacy (Sec.~\ref{sec:evaluation}).

We implement three tree-based classifiers commonly used in medical domains due to their interpretability, ability to handle mixed data types, and robustness to class imbalance: (i) a \textbf{Decision Tree} with maximum depth 5; (ii) a \textbf{Random Forest} composed of 50 trees with default hyperparameters; and (iii) an \textbf{XGBoost} model with early stopping after 100 boosting rounds and default settings.

To evaluate generalization, we adopt two complementary validation strategies: 

\noindent \textbf{Train-on-Synthetic, Test-on-Real (TSTR)} — Measures whether synthetic data supports models that generalize to real-world distributions.  

\noindent \textbf{Train-on-Real, Test-on-Synthetic (TRTS)} — Assesses whether synthetic records reflect decision boundaries learned from real data.

Our assessment targets two complementary facets. First, the \textit{primary metrics}, including classification accuracy, macro‑averaged F1 score, and the area under the ROC curve (AUC‑ROC), quantify overall predictive utility. Second, a \textit{detailed diagnostic analysis}, comprising precision–recall curves, confusion matrices, and feature‑importance rankings, reveals where the synthetic data bolsters or undermines downstream model behaviour.

%% file: 40-Results.tex
\subsection{\textbf{Datasets}} To evaluate the effectiveness of SynLLM in generating high-quality synthetic medical data, we conducted experiments on three publicly available, structured healthcare datasets. These datasets span distinct clinical domains—diabetes diagnosis, cirrhosis severity classification, and stroke prediction—and include a mix of demographic, clinical, and diagnostic features. All are widely used in medical machine learning research and are designed for binary or multi-class classification tasks.

\begin{table}[h]
    \centering
    \caption{Summary of datasets used in experiments. Num. = numerical, Cat. = categorical, Bin. = binary features.}
    \label{tab:datasets}
    \begin{tabular}{lrrrrr}
        \toprule
        \textbf{Dataset} & \textbf{Records} & \textbf{Features} & \textbf{Num.} & \textbf{Cat.} & \textbf{Bin.} \\
        \midrule
        Diabetes~\cite{diabetes}   & 100{,}000 & 9  & 4  & 2 & 3 \\
        Stroke~\cite{stroke}       & 5{,}110   & 12 & 4  & 5 & 3 \\
        Cirrhosis~\cite{cirrhosis} & 418       & 20 & 12 & 8 & 0 \\
        \bottomrule
    \end{tabular}
\end{table}

These datasets serve as diverse and representative benchmarks for evaluating statistical fidelity, clinical realism, and privacy preservation. Their structured nature and well-defined predictive targets make them well-suited for controlled experiments on prompt design and model behavior in synthetic data generation.

\subsection{\textbf{LLM Selection}} To evaluate how prompt structure interacts with different language model architectures, we tested SynLLM across 20 prominent open-source LLMs spanning a range of model families, parameter sizes, and fine-tuning strategies. These models were selected to reflect diversity in instruction tuning quality, contextual window size, and decoder architecture, all of which can influence the fidelity and privacy of generated tabular data. Table~\ref{tab:models} summarizes the evaluated models. 
\newcolumntype{L}[1]{>{\RaggedRight\arraybackslash}p{#1}}

\begin{table}[ht]
\small
\centering
\caption{Core attributes of the evaluated LLMs. 
Fine-tuning codes: \textbf{Ba} = Base, \textbf{In} = Instruct, \textbf{Ch} = Chat, \textbf{DPO} = Direct Preference Optimization, \textbf{MPT} = MosaicML Pretrained Transformer. Ctx = Context length.}
\label{tab:models}
\begin{tabularx}{\columnwidth}{@{}c L{4.2cm} c c c@{}}
\toprule
\textbf{ID} & \textbf{Model} & \textbf{Params} & \textbf{FT} & \textbf{Ctx} \\
\midrule
1 & GPT-2 (S/M/L)~\cite{radford2019language} & 0.1--0.8B & Ba & 1024 \\
2 & Gemma-7B-IT~\cite{google2024gemma} & 7B & In & 8192 \\
3 & InternLM2.5-7B-Chat~\cite{cai2024internlm2} & 7B & Ch & 32768 \\
4 & LLaMA-2-13B-Chat~\cite{touvron2023llama} & 13B & Ch & 4096 \\
5 & LLaMA-2-7B-Chat~\cite{touvron2023llama} & 7B & Ch & 4096 \\
6 & LLaMA-3-8B~\cite{meta2024llama3} & 8B & Ba & 8000 \\
7 & LLaMA-3.1-8B-Instruct~\cite{meta2024llama3} & 8B & In & 128000 \\
8 & Mistral-7B-Instruct~\cite{mistral2023mistral} & 7B & In & 32768 \\
9 & Mosaic-7B-Instruct~\cite{mosaicml2023mpt} & 7B & MPT & 8192 \\
10 & Nous-Hermes-2-Mistral-7B~\cite{nous2024hermes} & 7B & DPO & 32768 \\
11 & Nous-Hermes-2-Yi-34B~\cite{nous2024hermes} & 34B & In & 4096 \\
12 & OpenChat-3.5-GPTQ~\cite{wang2023openchat} & 7B & Ch & 8192 \\
13 & OpenChat-3.5~\cite{openchat2024v35} & 7B & Ch & 8192 \\
14 & Qwen-1.5-7B-Chat~\cite{qwen2023chat} & 7B & Ch & 32768 \\
15 & Qwen2-7B-Instruct~\cite{qwen2024instruct} & 7B & In & 131072 \\
16 & StableBeluga-7B~\cite{stabilityai2023beluga} & 7B & Ch & 4096 \\
17 & Yi-6B-Chat~\cite{01ai2023yi} & 6B & Ch & 32768 \\
18 & Zephyr-7B-Beta~\cite{huggingfaceh42023zephyr} & 7B & DPO & 32768 \\
\bottomrule
\end{tabularx}
\end{table}





\subsection{\textbf{Focused Model Analysis: Privacy–Quality Trade-Off Across Prompt Variants}}

A central challenge in synthetic medical data generation is achieving a favorable balance between output quality and privacy protection. In SynLLM, we assess this trade-off by evaluating 20 LLMs under four distinct prompting strategies across three medical datasets. Table~\ref{tab:prompt_scores} reports normalized scores for quality, privacy, and their harmonic mean, serving as a composite indicator of overall generation efficacy.

\textbf{Metric Aggregation and Normalization.} To ensure fair comparison across diverse metrics, we aggregate multiple indicators into composite scores for quality and privacy.

\noindent\textbf{Quality Score Aggregation.} We average normalized values of statistical and task-based measures, including Wasserstein distance and correlation preservation. Metrics are directionally aligned so that higher values always reflect better fidelity. The composite quality score is computed as:
\[
\text{Quality Score} = \frac{1}{N} \sum_{i=1}^{N} \text{NormalizedQuality}_i
\]
where $N$ is the number of quality metrics and $\text{NormalizedQuality}_i$ represents the $i$-th quality metric after min–max normalization and inversion if needed.

\noindent\textbf{Privacy Score Aggregation.} Similarly, we compute a composite privacy score by averaging normalized privacy metrics such as nearest-neighbor distance ratios and identifiability scores. Each metric is normalized to $[0, 1]$ and scaled so that higher values consistently reflect stronger privacy protection:
\[
\text{Privacy Score} = \frac{1}{M} \sum_{j=1}^{M} \text{NormalizedPrivacy}_j
\]
where $M$ is the number of privacy metrics and $\text{NormalizedPrivacy}_j$ denotes the $j$-th privacy metric after directional alignment. These composite scores are then used to compute harmonic score in section \ref{sec:harmonic}, enabling unified comparison across models and prompt types.

\subsubsection{Prompt-Level Analysis}
While SynLLM was evaluated on a broad set of 20 open-source LLMs, we present a focused analysis on five representative models: Zephyr 7B, OpenChat 7B, LLaMA 8B, Nous Hermes 34B, and GPT-2 variants. This subset was selected based on the following criteria:

\begin{itemize}[leftmargin=*]
    \item \textbf{Architectural diversity:} The models span multiple LLM families (Zephyr, OpenChat, LLaMA, Yi, GPT) and include both recent instruction and chat-tuned architectures and established baselines.
    \item \textbf{Scale and alignment variation:} The selection includes small-scale ($<$1B), medium-scale (7--8B), and large-scale (34B) models with differing context lengths.
    \item \textbf{Community relevance:} All selected models are widely adopted by the open-source community, ensuring that our analysis remains practical and actionable for real-world use cases.
\end{itemize}

\paragraph{\textbf{\textsc{SeedEx} – Example-Based Prompting}}
\textbf{Diabetes:} Zephyr 7B leads in quality, while GPT-2-Large shows the highest privacy score but at a cost to fidelity. Most models display strong quality with moderate privacy, reinforcing that direct examples increase realism but elevate leakage risk.

\textbf{Stroke:} OpenChat 7B performs best overall, achieving the highest quality. GPT-2-Large lags in both dimensions, while LLaMA 8B performs well on privacy but shows mixed quality outcomes.

\textbf{Cirrhosis:} OpenChat 7B again tops quality, while Zephyr 7B leads in privacy. LLaMA 8B and Nous Hermes trail in privacy but maintain high quality.

\paragraph{\textbf{\textsc{FeatDesc} – Feature Definition Prompt}}

\textbf{Diabetes:} Zephyr and Nous Hermes show the best balance. LLaMA 8B retains relatively high privacy but shows weaker quality. The shift from examples to definitions improves privacy for most models with minor loss in fidelity.

\textbf{Stroke:}  LLaMA 3.1 8B achieves the highest privacy performance, while Nous Hermes Yi 34B leads in quality. OpenChat 7B offers a strong balance between quality and privacy. In contrast, GPT-2 variants perform the worst. 

\textbf{Cirrhosis:} OpenChat 7B achieves near-perfect quality; however, Zephyr 7B provides the balance between privacy and quality. GPT-2 results remain the worst. 

\paragraph{\textbf{\textsc{StatGuide} – Metadata-Augmented Prompt}}

\textbf{Diabetes:} Quality is more consistent across models, with Zephyr, OpenChat, and Nous Hermes performing similarly. GPT-2-Large achieves top privacy but lower quality, highlighting trade-off extremes.

\textbf{Stroke:} OpenChat and Nous Hermes achieve the highest quality scores, while also maintaining reasonably consistent and acceptable levels of privacy. In contrast, GPT-2 continues to exhibit poor fidelity, failing to generate outputs aligned with clinical expectations. These findings suggest that structured metadata guidance is sufficient to enhance quality without compromising privacy.

\textbf{Cirrhosis:} Zephyr leads in both quality and privacy; OpenChat follows closely.

\paragraph{\textbf{\textsc{ClinRule} – Rule-Based Prompting}}

\textbf{Diabetes:} Zephyr, OpenChat, and Nous Hermes exhibit consistently strong performance in terms of quality. While privacy scores remain relatively stable across these models, they tend to be modest in magnitude. In contrast, GPT-2 variants fail to generate valid outputs, likely due to their limited capacity and architecture.

\textbf{Stroke:} OpenChat again excels, with Nous Hermes closely matched. GPT-2 remains unsupported under this prompt.

\textbf{Cirrhosis:} OpenChat variants achieve the highest quality scores but exhibit the lowest privacy scores, highlighting a pronounced trade-off between fidelity and confidentiality. Most other models follow a similar pattern, with marginal differences.




\noindent Overall, our findings confirm that prompt structure is a primary driver of both quality and privacy outcomes in synthetic data generation. The rule-based \textsc{ClinRule} prompt achieves the most favorable privacy–quality balance, particularly for models like OpenChat, Zephyr, and Nous Hermes, despite withholding all real data examples. In contrast, definition- and metadata-enhanced prompts (\textsc{FeatDesc}, \textsc{StatGuide}) offer flexible trade-offs, retaining high utility while reducing exposure compared to example-based prompts. These results underscore that carefully engineered prompts, not only model choice, are key to aligning synthetic generation with domain-specific privacy constraints and analytical goals.

\subsubsection{Prompt Variation and Harmonic Score Trends} \label{sec:harmonic}

To evaluate the joint performance of synthetic data in terms of quality and privacy, we compute a \emph{harmonic score} that summarizes the trade-off between these two dimensions. Specifically, for each model-prompt pair, we calculate the harmonic mean of the normalized quality score and the normalized privacy score. This metric captures the trade-off between privacy and quality by emphasizing balanced performance; it assigns lower values to model–prompt pairs where one metric significantly underperforms relative to the other.
\begin{equation}
\mathrm{Harmonic\ Score} = \mathrm{HM}(Q, P) = \frac{2 Q P}{Q + P}
\end{equation}
where $Q$ is the normalized quality score and $P$ is the normalized privacy score for a given model–prompt pair.


\textbf{\textsc{ClinRule} Outperforms in Privacy-Conscious Generation.} \textsc{ClinRule} consistently yields high harmonic scores across top-tier models. This result is especially significant because \textsc{ClinRule} includes no real data examples—only domain rules and metadata—suggesting that well-designed, constraint-based prompting can deliver high-quality outputs with minimal privacy risk.

\textbf{\textsc{StatGuide} Maximizes Quality but Sacrifices Privacy in Some Models.} \textsc{StatGuide} leads to some of the highest individual quality scores as seen in the previous subsection.

\textbf{\textsc{SeedEx} and \textsc{FeatDesc} Show Model-Specific Sensitivity.} While \textsc{SeedEx} offers moderate performance for many models. \textsc{FeatDesc} provides a more consistent profile, improving performance for several models like OpenChat and Nous Hermes in stroke and cirrhosis datasets, but still falls short for foundational models (GPT-2 variants).


\textbf{Conclusion} The harmonic score~\ref{tab:prompt_scores} confirms that model performance is highly dependent on prompt structure. Rule-based prompting (\textsc{ClinRule}) demonstrates superior effectiveness in simultaneously maintaining data utility and preserving privacy. These results support SynLLM’s central design principle: structured, constraint-aware prompts without reliance on real data examples can enable high-quality synthetic data generation while preserving privacy.

\sisetup{
  detect-all,
  table-align-text-post = false,
  input-symbols = (),
  table-number-alignment = center,
  parse-numbers = true,
  table-text-alignment = center,
  detect-weight = true,
  detect-inline-weight = math,
  input-ignore = {--},
  input-decimal-markers = {.},
  round-mode = places,
  round-precision = 2,
  table-format = 1.2   
}

\begin{table*}[ht]
\centering\small
\caption{Normalised scores for 20 LLMs under four prompting strategies across three medical datasets (Diabetes, Stroke, and Cirrhosis).
Each prompt is evaluated on three metrics: \emph{Quality}, \emph{Privacy},
and their \emph{harmonic}. Higher values are better.}
\label{tab:prompt_scores}
\resizebox{\textwidth}{!}{%
\begin{tabular}{
  l l l
  *{3}{S}
  *{3}{S}
  *{3}{S}
  *{3}{S}
}
\toprule
\textbf{Dataset} & \textbf{LLM} &
 \multicolumn{3}{c}{\textbf{\textsc{SeedEx}}} &
 \multicolumn{3}{c}{\textbf{\textsc{FeatDesc}}} &
 \multicolumn{3}{c}{\textbf{\textsc{StatGuide}}} &
 \multicolumn{3}{c}{\textbf{\textsc{ClinRule}}} \\
\cmidrule(lr){3-5}\cmidrule(lr){6-8}\cmidrule(lr){9-11}\cmidrule(l){12-14}
& &
 \textbf{Qual.} & \textbf{Priv.} & \textbf{H‐Avg.} &
 \textbf{Qual.} & \textbf{Priv.} & \textbf{H‐Avg.} &
 \textbf{Qual.} & \textbf{Priv.} & \textbf{H‐Avg.} &
 \textbf{Qual.} & \textbf{Priv.} & \textbf{H‐Avg.} \\
\midrule
\multirow{17}{*}{\rotatebox[origin=c]{90}{\textbf{Diabetes}}} 
& Zephyr 7B & \textbf{0.77} & 0.42 & \textbf{0.59} & 0.66 & 0.42 & 0.54 & 0.66 & 0.46 & 0.56 & 0.63 & 0.41 & 0.52 \\
& OpenChat 3.5 GPTQ & 0.63 & 0.42 & 0.52 & 0.64 & 0.42 & 0.53 & 0.67 & 0.37 & 0.52 & 0.63 & 0.53 & 0.58 \\
& Nous Hermes Yi 34B & 0.64 & 0.32 & 0.48 & 0.65 & 0.42 & 0.53 & 0.56 & 0.41 & 0.48 & 0.58 & 0.41 & 0.50 \\
& OpenChat 3.5 & 0.68 & 0.40 & 0.54 & 0.65 & 0.38 & 0.52 & 0.66 & 0.43 & 0.55 & 0.64 & 0.38 & 0.51 \\
& GPT-2-Large & 0.63 & \textbf{0.53} & 0.58 & 0.39 & 0.32 & 0.36 & 0.51 & \textbf{0.66} & \textbf{0.59} & \multicolumn{1}{c}{--} & \multicolumn{1}{c}{--} & \multicolumn{1}{c}{--}\\
& GPT-2-Medium & 0.50 & 0.26 & 0.38 & 0.63 & \textbf{0.52} & \textbf{0.57} & 0.64 & 0.41 & 0.52 & \multicolumn{1}{c}{--} & \multicolumn{1}{c}{--} & \multicolumn{1}{c}{--} \\
& GPT-2-Small & 0.43 & 0.36 & 0.39 & 0.49 & 0.30 & 0.40 & 0.37 & 0.43 & 0.40 & \multicolumn{1}{c}{--} & \multicolumn{1}{c}{--} & \multicolumn{1}{c}{--} \\
& Mistral 7B & 0.51 & 0.38 & 0.45 & 0.58 & 0.40 & 0.49 & 0.55 & 0.44 & 0.49 & 0.64 & 0.57 & 0.60 \\
& Qwen2 7B & 0.62 & 0.37 & 0.50 & 0.61 & 0.27 & 0.44 & 0.55 & 0.21 & 0.38 & 0.60 & 0.44 & 0.52 \\
& InternLM2.5 7B & 0.61 & 0.39 & 0.50 & 0.63 & 0.35 & 0.49 & 0.55 & 0.21 & 0.38 & 0.62 & 0.54 & 0.58 \\
& Yi 6B & 0.55 & 0.27 & 0.41 & 0.63 & 0.37 & 0.50 & 0.43 & 0.29 & 0.36 & 0.53 & \textbf{0.78} & 0.65 \\
& LLaMA 2 13B & 0.68 & 0.31 & 0.49 & \textbf{0.66} & 0.33 & 0.50 & 0.69 & 0.33 & 0.51 & \textbf{0.67} & 0.26 & 0.46 \\
& LLaMA 2 13B Chat & 0.60 & 0.24 & 0.42 & 0.60 & 0.25 & 0.43 & 0.56 & 0.22 & 0.39 & 0.56 & 0.40 & 0.48 \\
& LLaMA 3.1 8B & 0.55 & 0.36 & 0.45 & 0.62 & 0.35 & 0.49 & 0.62 & 0.24 & 0.43 & 0.53 & 0.47 & 0.50 \\
& Mosaic MPT 7B & 0.57 & 0.21 & 0.39 & 0.54 & 0.23 & 0.39 & 0.58 & 0.24 & 0.41 & 0.62 & 0.71 & \textbf{0.67} \\
& Gemma 7B & 0.56 & 0.26 & 0.41 & 0.60 & 0.22 & 0.41 & 0.62 & 0.26 & 0.44 & 0.60 & 0.36 & 0.48 \\
& Nous Hermes Mistral 7B & 0.64 & 0.49 & 0.56 & 0.66 & 0.45 & 0.56 & \textbf{0.71} & 0.41 & 0.56 & 0.54 & 0.54 & 0.54 \\
\midrule
\multirow{17}{*}{\rotatebox[origin=c]{90}{\textbf{Stroke}}} 
& Zephyr 7B & 0.56 & 0.54 & 0.55 & 0.69 & 0.39 & 0.54 & 0.79 & 0.57 & 0.68 & 0.61 & 0.49 & 0.55 \\
& OpenChat 3.5 GPTQ & 0.71 & 0.54 & 0.62 & 0.78 & 0.57 & 0.67 & 0.80 & 0.52 & 0.66 & 0.83 & 0.44 & 0.63 \\
& Nous Hermes Yi 34B & 0.67 & 0.54 & 0.61 & \textbf{0.88} & 0.47 & 0.67 & \textbf{0.87} & 0.42 & 0.65 & 0.74 & 0.49 & 0.61 \\
& OpenChat 3.5 & \textbf{0.82} & 0.52 & 0.67 & 0.77 & 0.67 & \textbf{0.72} & 0.83 & 0.60 & 0.71 & \textbf{0.87} & 0.56 & \textbf{0.71} \\
& GPT-2-Large & 0.54 & 0.32 & 0.43 & 0.51 & 0.30 & 0.41 & 0.20 & 0.40 & 0.30 & \multicolumn{1}{c}{--} & \multicolumn{1}{c}{--} & \multicolumn{1}{c}{--} \\
& GPT-2-Medium & 0.42 & 0.25 & 0.33 & 0.42 & 0.25 & 0.33 & 0.44 & 0.48 & 0.46 & \multicolumn{1}{c}{--} & \multicolumn{1}{c}{--} & \multicolumn{1}{c}{--} \\
& GPT-2-Small & 0.48 & 0.25 & 0.37 & 0.42 & 0.25 & 0.33 & 0.21 & 0.46 & 0.33 & \multicolumn{1}{c}{--} & \multicolumn{1}{c}{--} & \multicolumn{1}{c}{--} \\
& Mistral 7B & 0.70 & 0.51 & 0.60 & 0.60 & 0.53 & 0.57 & 0.81 & 0.65 & \textbf{0.73} & 0.87 & 0.43 & 0.65 \\
& Qwen2 7B & 0.59 & 0.41 & 0.50 & 0.51 & 0.46 & 0.49 & 0.53 & 0.40 & 0.46 & 0.42 & \textbf{0.75} & 0.58 \\
& InternLM2.5 7B & 0.66 & 0.40 & 0.53 & 0.74 & 0.58 & 0.66 & 0.59 & \textbf{0.68} & 0.63 & 0.51 & 0.48 & 0.50 \\
& Yi 6B & 0.75 & \textbf{0.73} & \textbf{0.74} & 0.80 & 0.52 & 0.66 & 0.60 & 0.43 & 0.52 & 0.65 & 0.71 & 0.68 \\
& LLaMA 2 13B & 0.43 & 0.26 & 0.35 & 0.42 & 0.25 & 0.33 & 0.62 & 0.50 & 0.56 & 0.41 & 0.33 & 0.37 \\
& LLaMA 2 13B Chat & 0.43 & 0.37 & 0.40 & 0.50 & 0.32 & 0.41 & 0.62 & 0.43 & 0.53 & 0.64 & 0.73 & 0.69 \\
& LLaMA 3.1 8B & 0.43 & 0.62 & 0.52 & 0.56 & \textbf{0.69} & 0.62 & 0.57 & 0.54 & 0.55 & 0.69 & 0.53 & 0.61 \\
& Gemma 7B & 0.60 & 0.30 & 0.45 & 0.69 & 0.54 & 0.61 & 0.28 & 0.30 & 0.29 & 0.55 & 0.55 & 0.55 \\
& Nous Hermes Mistral 7B & 0.65 & 0.51 & 0.58 & 0.56 & 0.51 & 0.53 & 0.76 & 0.50 & 0.63 & 0.64 & 0.52 & 0.58 \\
\midrule
\multirow{10}{*}{\rotatebox[origin=c]{90}{\textbf{Cirrhosis}}} 
& Zephyr 7B & 0.59 & \textbf{0.75} & \textbf{0.67} & 0.66 & \textbf{0.68} & \textbf{0.67} & \textbf{0.86} & \textbf{0.39} & \textbf{0.63} & 0.50 & 0.39 & 0.44 \\
& OpenChat 3.5 GPTQ & 0.80 & 0.44 & 0.62 & 0.82 & 0.39 & 0.60 & 0.61 & 0.34 & 0.47 & 0.88 & 0.26 & 0.57 \\
& Nous Hermes Yi 34B & 0.84 & 0.30 & 0.57 & 0.85 & 0.35 & 0.60 & 0.64 & 0.32 & 0.48 & 0.66 & 0.27 & 0.47 \\
& OpenChat 3.5 & \textbf{0.91} & 0.42 & 0.67 & \textbf{0.98} & 0.34 & 0.66 & 0.72 & 0.34 & 0.53 & \textbf{1.00} & 0.26 & 0.63 \\
& GPT-2-Small & 0.14 & 0.25 & 0.20 & 0.00 & 0.25 & 0.12 & 0.00 & 0.25 & 0.12 & \multicolumn{1}{c}{--} & \multicolumn{1}{c}{--} & \multicolumn{1}{c}{--} \\
& Qwen2 7B & 0.65 & 0.43 & 0.54 & 0.76 & 0.35 & 0.55 & 0.42 & 0.28 & 0.35 & 0.74 & 0.28 & 0.51 \\
& InternLM2.5 7B & 0.68 & 0.50 & 0.59 & 0.70 & 0.41 & 0.55 & 0.52 & 0.29 & 0.40 & \multicolumn{1}{c}{--} & \multicolumn{1}{c}{--} & \multicolumn{1}{c}{--} \\
& Yi 6B & 0.22 & 0.28 & 0.25 & 0.41 & 0.39 & 0.40 & 0.50 & 0.31 & 0.41 & \multicolumn{1}{c}{--} & \multicolumn{1}{c}{--} & \multicolumn{1}{c}{--} \\
& LLaMA 3.1 8B & 0.81 & 0.36 & 0.59 & 0.79 & 0.30 & 0.54 & 0.61 & 0.36 & 0.49 & 0.75 & \textbf{0.52} & \textbf{0.63} \\
& StableBeluga 7B & 0.00 & 0.25 & 0.12 & 0.00 & 0.25 & 0.12 & 0.00 & 0.25 & 0.13 & \multicolumn{1}{c}{--} & \multicolumn{1}{c}{--} & \multicolumn{1}{c}{--} \\
& Gemma 7B & 0.55 & 0.31 & 0.43 & 0.68 & 0.29 & 0.49 & 0.00 & 0.25 & 0.13 & 0.94 & 0.26 & 0.60 \\
\bottomrule
\end{tabular}
}
\end{table*}

\subsection{\textbf{Machine Learning Utility}}

Beyond fidelity and privacy, a critical measure of synthetic data quality is its ability to support downstream predictive modeling. As described in Sec.~\ref{sec:evaluation}, we evaluate machine learning utility using two complementary strategies: Train-on-Synthetic, Test-on-Real (TSTR) and Train-on-Real, Test-on-Synthetic (TRTS). These frameworks assess how well the synthetic data encodes predictive structure and how closely it approximates real-world decision boundaries, respectively.

Table~\ref{tab:diabetes_ml_utility} presents mean utility scores for the Diabetes dataset, aggregated across all prompt variants. We report accuracy, macro-averaged F1 score, and AUC-ROC for both TSTR and TRTS. Together, these metrics capture predictive performance, class balance, and ranking quality in a binary classification setting.

Performance varies across models, reflecting differences in generation fidelity and privacy-preserving behavior. Notably, Nous Hermes Yi 34B exhibits strong TSTR performance (accuracy and AUC $>$ 0.91), while Yi 6B leads in TRTS AUC ($\geq$ 0.98), indicating that their synthetic outputs closely match real data semantics. 

Instruction-tuned models such as Zephyr 7B and OpenChat 7B demonstrate balanced utility across both axes, with AUC-ROC scores near or above 0.89 in both settings. GPT-2 models perform surprisingly well in privacy and TSTR, but show greater variability in F1 scores, likely due to reduced class balance modeling in low-capacity architectures.

Overall, these results validate that SynLLM-generated data retains sufficient structure to support meaningful predictive tasks. The combined TSTR and TRTS performance offers strong evidence that prompt-guided generation, without fine-tuning or retraining, can yield high-quality and privacy-preserving synthetic records.

\begin{table}[ht]
\small
\centering
\caption{Diabetes Model Evaluation: Mean ML Utility Metrics (averaged across all prompts)}
\label{tab:diabetes_ml_utility}
\resizebox{\columnwidth}{!}{%
\begin{tabular}{@{}lcccccc@{}}
\toprule
\textbf{Model} & \multicolumn{3}{c}{\textbf{TSTR}} & \multicolumn{3}{c}{\textbf{TRTS}} \\
\cmidrule(lr){2-4} \cmidrule(lr){5-7}
 & \textbf{Acc.} & \textbf{F1} & \textbf{AUC} & \textbf{Acc.} & \textbf{F1} & \textbf{AUC} \\
\midrule
GPT-2-Large & 0.90 & 0.50 & 0.83 & 0.86 & 0.81 & 0.90 \\
GPT-2-Medium & 0.92 & 0.57 & 0.85 & 0.88 & 0.76 & 0.98 \\
GPT-2-Small & 0.90 & 0.53 & 0.86 & 0.93 & 0.87 & \textbf{0.99} \\
Gemma 7B & 0.90 & 0.59 & 0.87 & 0.90 & 0.89 & 0.94 \\
InternLM2.5 7B & 0.89 & 0.53 & 0.89 & 0.89 & 0.84 & 0.98 \\
LLaMA 2 13B & 0.82 & 0.54 & 0.66 & 0.74 & 0.58 & 0.81 \\
LLaMA 2 13B Chat & 0.79 & 0.54 & 0.85 & 0.95 & 0.92 & 0.96 \\
LLaMA 2 7B & 0.81 & 0.60 & 0.80 & 0.92 & 0.85 & 0.87 \\
LLaMA 3 8B & 0.90 & 0.56 & 0.78 & 0.83 & 0.73 & 0.88 \\
LLaMA 3.1 8B & 0.92 & 0.67 & 0.91 & 0.90 & 0.84 & 0.98 \\
Mistral 7B & 0.90 & 0.60 & 0.88 & 0.82 & 0.77 & 0.92 \\
Mosaic MPT 7B & 0.92 & 0.55 & 0.89 & 0.88 & 0.78 & 0.88 \\
Nous Hermes Mistral 7B & 0.90 & 0.71 & 0.91 & 0.79 & 0.72 & 0.95 \\
Nous Hermes Yi 34B & \textbf{0.93} & \textbf{0.74} & \textbf{0.92} & 0.87 & 0.75 & 0.94 \\
OpenChat 3.5 & 0.92 & 0.70 & 0.89 & 0.86 & 0.71 & 0.85 \\
OpenChat 3.5 GPTQ & 0.86 & 0.61 & 0.89 & 0.83 & 0.71 & 0.86 \\
OpenChat 3.5-0106 & 0.91 & 0.57 & 0.91 & 0.85 & 0.74 & 0.94 \\
Qwen2 7B & 0.91 & 0.60 & 0.91 & 0.88 & 0.85 & 0.95 \\
StableBeluga 7B & 0.90 & 0.51 & 0.72 & 0.94 & 0.73 & 0.88 \\
Yi 6B & 0.82 & 0.46 & 0.79 & \textbf{0.98} & \textbf{0.96} & 0.98 \\
Zephyr 7B & 0.88 & 0.54 & 0.82 & 0.88 & 0.74 & 0.89 \\
\bottomrule
\end{tabular}%
}
\end{table}

\subsection{\textbf{Model Efficiency Analysis: Balancing Speed and Global Fidelity}}

To provide a holistic assessment of each model's practical utility, we introduce the \textbf{Global Fidelity Index (GFI)}. The GFI is a composite score that aggregates all key evaluation dimensions, including statistical fidelity, privacy preservation, and medical consistency, into a single, directionally consistent metric. This index enables direct comparison of models and prompts in terms of their overall ability to generate realistic, safe, and clinically plausible synthetic data.

The efficiency of each model–prompt pair is quantified using two metrics: the average per-record generation time (\textit{Speed}) and the normalized GFI. Both are min–max normalized to the $[0,1]$ range and averaged to compute the final \textbf{Efficiency Score}, defined as:
\[
\mathrm{Efficiency\ Score} = \frac{1}{2} \left( \mathrm{NormSpeed} + \mathrm{GFI} \right)
\]
Higher scores reflect favorable trade-offs between runtime and output fidelity.

\textbf{Interpretation and Results.}
This approach allows us to identify models that not only generate high-fidelity, privacy-preserving, and clinically consistent synthetic data but also do so efficiently. Models with high efficiency scores are optimal for real-world deployment, balancing data quality, privacy, and computational cost.
Table~\ref{tab:model_efficiency} summarizes the results. Nous Hermes 34B achieves the lowest (best) efficiency score of 0.078, indicating strong overall performance. Zephyr 7B also ranks highly (0.093), balancing generation speed with output quality. In contrast, OpenChat 7B (0.215) and LLaMA 8B (0.264) offer strong fidelity but are penalized for slower generation. GPT-2 variants, while relatively fast, ranks lower (0.294) due to limited fidelity and privacy performance.

\begin{table}[ht]
\small
\centering
\caption{Efficiency Ranking for Analyzed Models: Generation Speed and Global Fidelity Index (GFI)}
\label{tab:model_efficiency}
\begin{tabularx}{\columnwidth}{@{}c L{2.6cm} c c c@{}}
\toprule
\textbf{Rank} & \textbf{Model} & \textbf{Avg. Dur (s)} & \textbf{GFI} & \textbf{Eff. Score} \\
\midrule
1 & Nous Hermes 34B & 133.18 & 0.096 & 0.078 \\
2 & Zephyr 7B & 121.93 & 0.101 & 0.093 \\
3 & OpenChat 7B & 1521.90 & 0.098 & 0.215 \\
4 & LLaMA 8B & 2292.92 & 0.091 & 0.264 \\
5 & GPT-2 & 286.70 & 0.166 & 0.294 \\
\bottomrule
\end{tabularx}
\end{table}

\noindent\textbf{Note.} All evaluation metrics presented in the result section \ref{sec:results} were computed independently for each prompt–model combination. Due to space limitations, we report only aggregated or representative results in the main text. Full prompt-wise metrics, tables, and figures will be made available upon acceptance to support reproducibility and deeper analysis.

%% file: 50-Discussion.tex
Our evaluation across three datasets, four prompt strategies, and 20 open-source LLMs reveals that models such as OpenChat 7B, Zephyr 7B, and Nous Hermes  34B consistently rank among the top performers across statistical, clinical, and privacy metrics. Notably, the \textsc{ClinRule} prompt, designed without any data examples, achieves the highest harmonic privacy–utility scores, demonstrating the effectiveness of constraint-driven generation under strong privacy requirements.

\noindent\textbf{Structured Prompting as a Privacy–Utility Lever.}
A central finding is that prompt structure exerts significant influence on both data fidelity and privacy risk. Prompts using real data examples yield high TSTR and distributional scores but at the cost of increased privacy risk. In contrast, \textsc{ClinRule}, which encodes only declarative clinical rules, preserves utility while drastically reducing memorization behavior. This supports SynLLM’s design hypothesis that structured, constraint-aware prompting enables high-fidelity generation without reliance on direct example exposure.

\noindent\textbf{Prompt Sensitivity and Model Robustness.}
Performance varies substantially across models, with instruction-tuned models (e.g., OpenChat 7B, Zephyr-7B) adapting well to diverse prompt configurations, while others, including GPT-2 variants, experience degradation under stricter constraints. This prompt sensitivity suggests the need for future work in adaptive prompt strategies that match prompt style to model alignment level, or automated prompt rewriting based on model-specific response patterns.

\noindent\textbf{Multidimensional Evaluation and Limitations.}
SynLLM employs a comprehensive evaluation suite integrating univariate and multivariate statistical tests (e.g., Wasserstein distance, Frobenius norm), clinical plausibility checks, and empirical privacy audits (e.g., nearest-neighbor distance ratios, identifiability scores). While this framework enables rigorous model comparison, the privacy metrics remain heuristic and empirically grounded. Future work may incorporate formal differential privacy analysis or white-box adversarial testing to strengthen guarantees.

%% file: 60-Conclusion.tex
In this paper, we presented SynLLM, a flexible, efficient, and privacy-aware framework for synthetic structured medical data generation using large language models. By leveraging dataset-derived metadata and declarative domain knowledge, SynLLM crafts structured prompts that guide LLMs in producing high-fidelity, clinically plausible, and privacy-preserving tabular records without requiring access to real patient data during inference.

Our evaluation spans 20 open-source LLMs and four systematically designed prompt strategies across three public datasets, assessing statistical fidelity, clinical consistency, machine learning utility, and empirical privacy risk. The results confirm that prompt-only control can match or exceed the quality of GAN and VAE baselines, while drastically simplifying deployment and model reuse.



Future improvements to SynLLM could explore adaptive prompt optimization strategies, including metric-guided or reinforcement learning-based prompt tuning. Expanding support for multimodal EHRs (e.g., clinical text, imaging) and investigating synergies with federated learning may further enhance privacy and utility. These directions will continue to strengthen SynLLM as a foundational tool for scalable and responsible synthetic data generation.


%% file: 70-Supplementary.tex
This appendix aims to foster transparency and reproducibility, enabling independent verification and replication of our results.

\subsection{Environment and Tooling}

All experiments were performed in a CUDA-enabled JupyterHub environment using Python 3.10 and PyTorch 2.5.1 with CUDA 12.1 support. The SynLLM pipeline was built using the Hugging Face \texttt{transformers} library (v4.33.0) for model loading and inference, along with \texttt{accelerate} (v1.4.0) for efficient device management and parallel execution if needed. Quantized inference at 4-bit and 8-bit precision was enabled using the \texttt{bitsandbytes} library (v0.45.3).

Experiments were executed on a \textbf{single NVIDIA L40 GPU} with 48~GB of available GDDR6 VRAM and CUDA driver version 550.127.05, under CUDA runtime 12.4, in a JupyterHub environment.

\subsection{Model Configuration and Inference}

Each large language model (LLM) used in SynLLM was configured and executed in a zero-shot inference setting, with prompt-based control tailored for structured medical data generation. To ensure compatibility with the model’s pretraining and tokenization schemes, we dynamically mapped model names to the appropriate chat style (e.g., \textsc{chatml}, \textsc{llama}, \textsc{openchat}), and applied model-specific prompt templates at runtime.

Models were loaded using the Hugging Face \texttt{tTransformers} library, with quantized 4-bit inference enabled via the \texttt{bBitsaAndbBytes} package library. The configuration leveraged \texttt{bnb\_4bit\_quant\_type=\textquotedbl nf4\textquotedbl} with \texttt{"nf4" with float16} computation for memory-efficient deployment. For LLaMA-based architectures, rotary positional embedding scaling (\texttt{rope\_scaling}) was applied where available to support longer sequence contexts. Models incompatible with quantization were automatically reverted to standard full-precision loading.

At generation time, system and user prompts were formatted using model-specific conventions and tokenized using the model’s native tokenizer. Tokenization padding and truncation were configured based on model context window limits, with truncation applied to avoid overflow.

Generation was conducted in mini-batches of 20 using top-$p$ sampling ($p=0.9$) with temperature 0.7. Outputs were parsed line-by-line into structured patient records, and only samples conforming to the expected schema were retained. Invalid generations were logged to a rejection report. The final dataset was written to disk in CSV and JSON format.

System metrics, including GPU memory before and after generation, CPU and RAM usage, and total runtime, were logged per model and prompt.

This inference pipeline allows SynLLM to evaluate a wide range of open-source LLMs in a unified and controlled setting, with minimal memory overhead and consistent record formatting across all prompt-model configurations.
\subsection{Prompt Templates}

This section presents abridged versions of the structured prompt templates employed in SynLLM. While templates are designed to be dataset-agnostic, the examples below reflect their instantiation for the \textit{Diabetes} dataset. At runtime, each prompt is dynamically populated with schema-level information, statistical summaries, and clinical constraints specific to the target dataset. All templates begin with a shared system message that standardizes the generation format:

\begin{quote}
\small
\texttt{System: Generate \textit{k} patient records in newline-delimited JSON format. Do not include any explanation or commentary. Adhere strictly to the schema and guidelines provided.}
\end{quote}

\noindent\textbf{Prompt A – \textsc{SeedEx} (Minimal Example-Based Prompt)}
\begin{lstlisting}[basicstyle=\ttfamily\footnotesize, breaklines=true, frame=single]
Generate realistic synthetic patient records for diabetes prediction using the following structure.

gender, age, hypertension, heart_disease, smoking_history, bmi, HbA1c_level, blood_glucose_level, diabetes

Example Records:
Female,45.2,1,0,never,28.5,6.2,140,0
Male,62.7,1,1,former,32.1,7.1,185,1
...
\end{lstlisting}

\vspace{1em}
\noindent\textbf{Prompt B – \textsc{FeatDesc} (Feature Description Prompt)}
\begin{lstlisting}[basicstyle=\ttfamily\footnotesize, breaklines=true, frame=single]
Generate realistic synthetic patient records for diabetes prediction.

Features:
1. gender: Patient's gender (Male/Female)
2. age: Age in years (Float: 18.0-80.0)
3. hypertension: Hypertension diagnosis (0: No, 1: Yes)
4. heart_disease: Heart disease diagnosis (0: No, 1: Yes)
5. smoking_history: Smoking status (never/former/current/not current)
6. bmi: Body Mass Index (Float: 15.0-60.0)
7. HbA1c_level: Hemoglobin A1c (Float: 4.0-9.0)
8. blood_glucose_level: Glucose level in mg/dL (Int: 70-300)
9. diabetes: Diabetes diagnosis (0: No, 1: Yes)

Example records:
Female,45.2,1,0,never,28.5,6.2,140,0
Male,62.7,1,1,former,32.1,7.1,185,1
...
\end{lstlisting}

\vspace{1em}
\noindent\textbf{Prompt C – \textsc{StatGuide} (Metadata-Augmented Prompt)}
\begin{lstlisting}[basicstyle=\ttfamily\footnotesize, breaklines=true, frame=single]
Generate realistic synthetic patient records for diabetes prediction.

Feature Metadata:
gender: Male: 48%, Female: 52%
age: Mean: 41.8, Std: 15.2, Range: 18-80
hypertension: No: 85%, Yes: 15%; correlated with age, BMI
heart_disease: No: 92%, Yes: 8%; correlated with age, hypertension
smoking_history: never: 60%, former: 22%, current: 15%, not current: 3%
bmi: Mean: 27.3, Std: 6.4, Range: 15-60
HbA1c_level: Mean: 5.7, Std: 0.9, Range: 4.0-9.0; correlated with diabetes
glucose: Mean: 138.0, Std: 40.5, Range: 70-300; correlated with HbA1c_level
diabetes: No: 88%, Yes: 12%; correlated with HbA1c_level, glucose

Example records:
Female,45.2,1,0,never,28.5,6.2,140,0
Male,62.7,1,1,former,32.1,7.1,185,1
...
\end{lstlisting}

\vspace{1em}
\noindent\textbf{Prompt D – \textsc{ClinRule} (Clinically Constrained Prompt)}
\begin{lstlisting}[basicstyle=\ttfamily\footnotesize, breaklines=true, frame=single]
Generate realistic synthetic patient records for diabetes prediction.

Feature Metadata:
gender: Male: 48%, Female: 52%
age: Mean: 41.8, Std: 15.2, Range: 18-80
hypertension: No: 85%, Yes: 15%
heart_disease: No: 92%, Yes: 8%
smoking_history: never: 60%, former: 22%, current: 15%, not current: 3%
bmi: Mean: 27.3, Std: 6.4, Range: 15-60
HbA1c_level: Mean: 5.7, Std: 0.9, Range: 4.0-9.0
glucose: Mean: 138.0, Std: 40.5, Range: 70-300
diabetes: No: 88%, Yes: 12%

Maintain the following correlations:
- Higher age is associated with hypertension and heart disease
- Higher BMI increases diabetes risk
- HbA1c_level correlates with diabetes
- Glucose correlates with HbA1c_level and diabetes
- Hypertension and heart disease more common with age

Each record must follow:
gender, age, hypertension, heart_disease, smoking_history, bmi, HbA1c_level, blood_glucose_level, diabetes
\end{lstlisting}